\documentclass{bmvc2k}

\title{IPGPhormer: Interpretable Pathology Graph-Transformer for Survival Analysis}

\usepackage{marvosym}
\usepackage{amssymb}
\addauthor{Guo Tang}{220810318@stu.hit.edu.cn}{1 $*$}
\addauthor{Songhan Jiang}{23S151019@stu.hit.edu.cn}{1 $*$}
\addauthor{Jinpeng Lu}{jinpenglu231@gmail.com}{1,2}
\addauthor{Linghan Cai}{cailh@stu.hit.edu.cn}{1}
\addauthor{Yongbing Zhang}{ ybzhang08@hit.edu.cn}{1 \Letter}

\addinstitution{
 Harbin Institute of Technology, Shenzhen\\
 Shenzhen, China
}
\addinstitution{
 University of Science and Technology of China\\
 Hefei, China
}

\runninghead{Tang ET AL.}{IPGPhormer: Interpretable Survival Analysis}

\usepackage{booktabs}
\usepackage{amsmath}

\usepackage{bbding}
\usepackage{colortbl}
\usepackage{multirow, array, xcolor}
\usepackage[T1]{fontenc}
\usepackage{graphicx,verbatim}
\usepackage{booktabs} % 用于更好的表格线条
\usepackage{footnote}
\usepackage{pifont} % 用于\Checkmark和\XSolidBrush符号
\usepackage{float}      % 提供[H]浮动定位
\usepackage{tabularx}   % 表格支持

\begin{document}

\maketitle

\begin{abstract}
% This document demonstrates the format requirements for papers submitted
% to the British Machine Vision Conference.  The format is designed for
% easy on-screen reading, and to print well at one or two pages per sheet.
% Additional features include: pop-up annotations for
% citations~\cite{Authors06,Mermin89}; a margin ruler for reviewing; and a
% greatly simplified way of entering multiple authors and institutions.

% {\bf All authors are encouraged to read this document}, even if you have
% written many papers before.  As well as a description of the format, the
% document contains many instructions relating to formatting problems and
% errors that are common even in the work of authors who {\em have}
% written many papers before.
Pathological images play an essential role in cancer prognosis, while survival analysis, which integrates computational techniques, can predict critical clinical events such as patient mortality or disease recurrence from whole-slide images (WSIs). 
Recent advancements in multiple instance learning have significantly improved the efficiency of survival analysis. However, existing methods often struggle to balance the modeling of long-range spatial relationships with local contextual dependencies and typically lack inherent interpretability, limiting their clinical utility.
To address these challenges, we propose the Interpretable Pathology Graph-Transformer (IPGPhormer), a novel framework that captures the characteristics of the tumor microenvironment and models their spatial dependencies across the tissue. IPGPhormer uniquely provides interpretability at both tissue and cellular levels without requiring post-hoc manual annotations, enabling detailed analyses of individual WSIs and cross-cohort assessments. 
Comprehensive evaluations on four public benchmark datasets demonstrate that IPGPhormer outperforms state-of-the-art methods in both predictive accuracy and interpretability. 
% By offering a robust and precise solution for cancer prognosis assessment, IPGPhormer bridges the gap between computational methods and clinical practice, paving the way for more reliable and interpretable decision-support tools in pathology.
In summary, our method, IPGPhormer, offers a promising tool for cancer prognosis assessment, paving the way for more reliable and interpretable decision-support systems in pathology. The code is publicly available at \href{https://anonymous.4open.science/r/IPGPhormer-6EEB}{https://anonymous.4open.science/r/IPGPhormer-6EEB}.
\end{abstract}

%-------------------------------------------------------------------------
\section{Introduction}
\label{sec:intro}
% From 2009, the proceedings of BMVC (the British Machine Vision
% Conference) will be published only in electronic form.  This document
% illustrates the required paper format, and includes guidelines on
% preparation of submissions.  Papers which fail to adhere to these
% requirements may be rejected at any stage in the review process.

% \LaTeX\ users should use this template in order to prepare their paper.
% Users of other packages should emulate the style and layout of this
% example.  Note that best results will be achieved using {\tt pdflatex},
% which is available in most modern distributions.

Computational pathology integrates pathology and computer science to develop methods for analyzing pathological images, providing valuable insights for cancer diagnosis and prognosis. A key task in this field is survival analysis, which predicts the time from a known origin to critical events such as death or disease recurrence~\cite{aalen2008survival,dey2022survival}. Previous works generally rely on expert annotations to delineate cancer regions from ultrahigh-resolution whole slide images (WSIs) for subsequent analysis~\cite{collins2015new,jackson2020single}; however, this process is time-consuming and labor-intensive, heavily depending on the pathologists.

Recent advances in deep learning, particularly in multiple instance learning (MIL) algorithms, have significantly improved the efficiency of survival analysis. These advancements have shifted the focus from analyzing local regions of interest to processing large-scale gigapixel WSIs, unlocking the potential to better understand disease progression~\cite{liu2022deep,liu2024advmil}. 
Currently, some MIL frameworks use attention-based mechanisms to highlight patch interactions. For instance, ABMIL~\cite{ilse2018attention} and CLAM~\cite{lu2021data} employ a local attention mechanism to locate key patches, while TransMIL~\cite{shao2021transmil} and Surformer~\cite{Surformer} leverage Transformer~\cite{vaswani2017attention} to model long-range dependencies. SurvTRACE~\cite{wang2022survtrace} uses an adaptive Transformer to model the density function of nonparametric survival analysis, but its multi-tasking framework limits the overall performance. However, more importantly, these methods still struggle to characterize the complexity of tissue microstructures and spatial dependencies due to the highly heterogeneous nature of pathological tissues and the lack of explicit modeling of biologically relevant spatial interactions.

In contrast, graph-based methods directly model contextual relationships, offering a promising approach to characterizing complex pathological topology structures. 
For instance, Patch-GCN~\cite{chen2021whole} pioneers the use of graph convolutional networks for survival analysis, while WiKG~\cite{li2024dynamic} constructs feature similarity-driven graphs by knowledge-guided attention mechanisms to capture inter-patch semantic relationships. GRASP~\cite{mirabadi2024grasp} proposes a lightweight multi-scale graph framework to aggregate the node features in pathological images through convergent nodes. However, these methods are similar to most attention-based MIL approaches, utilizing attention mechanisms primarily for interpretability.
More recent approaches attempt to further enhance interpretability. TEA~\cite{TEA} employs the Integrated Gradients algorithm to attribute risk to each patch. GTP~\cite{zheng2022graph} introduces GraphCAM, which leverages gradients and propagation within the network. Additionally, HEAT~\cite{chan2023histopathology} utilizes Granger causality localization~\cite{lin2021generative} to generate causality heatmaps.
Despite these advancements,
existing graph-based methods focus on computing slide-level risk scores rather than directly assessing patch-level risks.
Moreover, while they can effectively capture spatial and semantic features, they still rely on post-hoc techniques to derive interpretability, which limits their transparency and clinical utility. 
This reliance on post-hoc analysis undermines the critical need for pathologist-friendly reasoning and accountability in clinical settings, which are essential for building trust and ensuring reliability in routine workflows.

To overcome these challenges, we propose the Interpretable Pathology Graph-Transformer (IPGPhormer) framework. This explainable model enables effective measures of patch risk scores.
Our approach constructs a multi-scale graph, utilizing heterogeneous graphs at high magnification to capture microenvironment subtype features and homogeneous tissue graphs at low magnification to characterize morphological features. 
Additionally, we design separate Patch-Level and Region-Level Feature Transfer Modules to better model tumor microenvironment morphology and long-range spatial dependencies. 
Unlike existing methods that rely on post-hoc manual annotations, IPGPhormer provides inherent interpretability by predicting patch-level risks and statistical analyses to achieve tissue- and cell-level interpretability grounded in pathology.
Our main contributions are as follows:
(1) We introduce a novel framework that effectively captures local spatial awareness and long-range dependencies, enabling precise patch-level risk prediction.
(2) We propose an interpretable method that eliminates the need for post-hoc manual annotations. This method supports tissue-level interpretability for individual WSIs and cell-level interpretability across cohorts.
(3) Extensive evaluations on four public benchmark datasets demonstrate that our framework outperforms state-of-the-art methods in accuracy and interpretability, highlighting its potential as a reliable decision tool for cancer prognosis assessment.
% SDFdzfdgzdsfgdszfszKSJDBFkjDBSKJFbkDJBFkjDBKFjbDKJBFkjDBFSDFkjhjkhkjhdskjhfkjsdhfkjsdhkfjhsdkjfhsdkjh, dkjfhsdjkfhsSDKJFHkjd

\section{Methodology}

\begin{figure}[t]
\includegraphics[width=\textwidth]{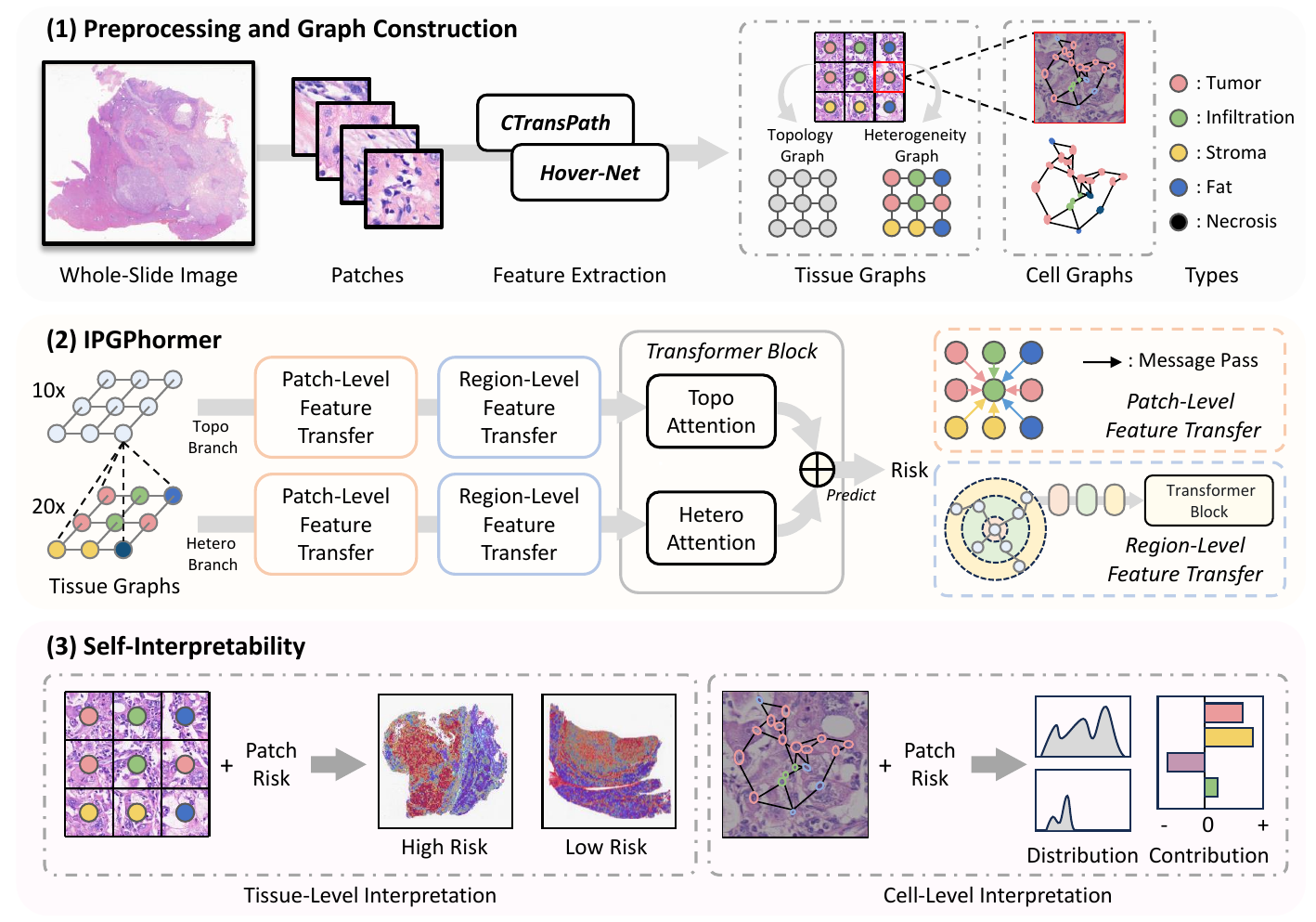}
\caption{
Overview of the proposed IPGPhormer architecture. 
First, we use HoverNet to construct both tissue graphs and cell graphs, and then extract patch features using CTransPath.
The Patch-Level Transfer Module leverages GAT for local spatial awareness, while the Region-Level Feature Transfer Module converts the graph data into a sequential embedding format and feeds it into the Transformer blocks to capture long-range dependencies. 
Finally, the patch risk score establishes a link between model outputs and both tissue-level and cell-level interpretability.
} 
\label{fig1}
\end{figure}

The overview of our proposed framework IPGPhormer is given in Fig. \ref{fig1}. The following sections provide detailed explanations of the method.

\subsection{Graph Construction}\label{sec:graph_construction}
We first divide the whole slide into patches of 256$\times$256, following CLAM~\cite{lu2021data}, and then perform cell segmentation and classification on each patch using a HoverNet~\cite{graham2019hover} pretrained on the PanNuke dataset~\cite{gamper2019pannuke}.
% We first apply Otsu’s thresholding algorithm~\cite{chen2021whole} to the whole slide to distinguish foreground from background, then divide it into 256×256 patches following CLAM~\cite{lu2021data}, and finally perform cell segmentation and classification on each patch using a HoverNet~\cite{graham2019hover} pretrained on the PanNuke dataset~\cite{gamper2019pannuke}. 
The cell detection results output by Hovernet are used for subsequent cell-level statistical feature extraction. Based on the cell classification results, we determine the patch type through majority voting of nucleus types and finally divide it into five categories. Next, we construct tissue graphs using patches as nodes and extract patch embeddings ($d=768$) at both magnifications using the CTransPath~\cite{wang2022transformer}. 
Then, we construct a heterogeneous graph at 20$\times$ magnification and a homogeneous graph at 10$\times$ magnification. This separation is motivated by the complementary information available at different scales: high magnifications reveal fine-grained, cellular, and micro-environment-level features that benefit from heterogeneous modeling, whereas lower magnifications capture global tissue architecture and contextual patterns better represented with homogeneous graphs; jointly modeling both scales enables the network to leverage local cellular detail and global structural context. 
We take the Euclidean distance in the spatial domain as the measure and compute the adjacency matrix to capture node connectivity using the K-Nearest Neighbor algorithm (K-NN). This results in a multi-scale tissue graph $\mathcal{G} \equiv (\mathcal{G}_L, \mathcal{G}_H) = (\mathcal{V}_L, \mathcal{E}_L, \mathcal{V}_H, \mathcal{E}_H, \mathcal{T}_H, \mathcal{E}_{L \leftrightarrow H})$, where $\mathcal{G}_L$ and $\mathcal{G}_H$ correspond to the low- and high-magnification graphs. The sets of nodes and edges are $\mathcal{V}_i$ and $\mathcal{E}_i$, with $\mathcal{T}_H$ representing node types in the high-magnification graph. 
% For $v \in \mathcal{V}_H$, the entity type is given by $\tau(v) \in \mathcal{A}_H$. 
The set $\mathcal{E}_{L\leftrightarrow H}$ denotes edges between low- and high-magnification nodes, constructed through a cross-scale alignment method performed via spatial coordinate projection.
% Each node $v$ has a $d$-dimensional feature $x_v$.
The normalized adjacency matrix is defined as $\tilde{\mathcal{A}}=\tilde{\mathcal{D}}^{-1/2}{\mathcal{A}}\tilde{\mathcal{D}}^{-1/2}$, $\tilde{\mathcal{D}}$ denotes the corresponding degree matrix, $\mathcal{A}$ denotes the adjacency matrix.

\subsection{Feature Transfer Module}
\noindent \textbf{Patch-Level Feature Transfer Module:}
We employ a Graph Attention Network (GAT)~\cite{velivckovic2017graph} with $L=3$ layers to capture the local spatial relationships and heterogeneity within $L$-hop neighborhoods. The feature updating process is formulated as:
% For node $v$, we set $L=3$ iteratively aggregate neighborhood information over $L$ steps, which captures local spatial relationships and heterogeneity in its $L$-hop ego-graph.
% Updating the feature matrix for high and low scales will be represented as the same $\mathbf{H}^{(L)}$ as the module outputs for simplicity.
% Both the topology and heterogeneity graphs utilize GAT~\cite{velivckovic2017graph} due to its simplicity and effectiveness, as formulated:
\begin{equation}
% \small % \vec
\alpha_{vu}^{(l)} = \frac{\exp\left( \mathrm{LeakyReLU}\left( \vec{\mathbf{a}}^T \left[ W {h}_v^{(l)}, W {h}_u^{(l)} \right] \right) \right)}{\sum_{k \in \mathcal{N}_v} \exp\left( \mathrm{LeakyReLU}\left( \vec{\mathbf{a}}^T \left[ W {h}_v^{(l)}, W {h}_k^{(l)} \right] \right) \right)},
\end{equation}

\begin{equation}
{h}_v^{(l+1)} = \mathrm{ReLU}\left( \sum_{u \in \mathcal{N}_v} \alpha_{vu}^{(l)} {W} {h}_u^{(l)} \right),
\end{equation}
where ${h}_v^{(l)}$ denotes the feature vector of node $v$ at the $l$-th layer, $\alpha_{vu}^{(l)}$ represents the attention weight between nodes $v$ and $u$, $\mathcal{N}_v$ is the set of neighboring nodes of $v$, and $[\cdot,\cdot]$ denotes vector concatenation. The learnable parameters include $W\in\mathbb{R}^{d' \times d}$ and $\vec{\mathbf{a}}\in\mathbb{R}^{2d'}$. For simplicity, we denote the final node representations after $L$ iterations as $\mathbf{H}^{(L)}$, which serves as the module output for both high and low scales.
% where $W\in\mathbb{R}^{d' \times d}$ and $\vec{\mathbf{a}}\in\mathbb{R}^{2d'}$ are learnable parameters, $[\cdot, \cdot]$ denotes the concatenation,
% $\mathcal{N}_v$ is the neighbor set of node $v$,
% ${h}_v^{(l)}$ represents the feature vector of node $v$ at layer $l$ and $\mathbf{H}^{(L)}$ represents the concatenated ${h}_v^{(l)}$. 

The module handles two distinct branches: Topology Graph and Heterogeneous Graph, each with its specific operations.
Each node $v$ is associated with an original $d$-dimensional feature vector $x_v$, collectively denoted by the graph feature matrix $\mathbf{X}_H$ and $\mathbf{X}_L$ of high- and low-scales.
% In the Topology Information Encoder (TIE) process
In the Topology Branch, we set the initial node representation as the input $\mathbf{H}^{(0)}=\mathbf{X}_L$ and name the module Topology Information Encoder (TIE).
% In the Heterogeneous Information Encoder (HIE) process
In the Heterogeneous Branch, to capture heterogeneity, we initialize the heterogeneous encoding of node $v$ by concatenating its type vector $\tau_v$ and feature vector $r_v = [\tau_v,x_v]$, where $\tau_v$ is the one-hot type embedding. The input is defined as $\mathbf{H}^{(0)} = \mathbf{R}$, where $\mathbf{R}$ represents the concatenated heterogeneous node encoding collectively denoted by $r_v$. We name the module Heterogeneous Information Encoder (HIE).

\noindent \textbf{Region-Level Feature Transfer Module:} 
After obtaining the feature matrix $\mathbf{H}$ from the Patch-Level Transfer Module, we convert the graph into a sequence to serve as input to the transformer block, as follows:
\begin{equation}
% \small
\mathbf{Z} = \sum_{s=0}^S \beta_s \mathbf{Z}^{(s)}, \quad \mathbf{Z}^{(s)} = \tilde{\mathcal{A}} \mathbf{Z}^{(s-1)}, \quad \mathbf{Z}^{(0)} = \mathbf{H},
\end{equation}
where $\tilde{\mathcal{A}}$ denotes the normalised adjacency matrix, $\beta_s$ is the learnable coefficient for aggregation at step $s$, $S$ is a hyperparameter set to 3, and $\mathbf{Z}^{(s)}$ represents the hidden node representations at the $s$-th propagation step. This process results in the updated node representations $\mathbf{Z}$.
The decoupled design improves computational efficiency while enabling the model to capture structural information.

\subsection{Graph-Based Transformer Block}
In this module, we employ a global self-attention mechanism to compute single-scale region-level attention on the graphs $\mathcal{G}_{L}$ and $\mathcal{G}_{H}$. Subsequently, we fuse these attention maps into a unified attention map. 
The matrices are defined as $\mathbf{Q} = W_q \mathbf{Z}$, $\mathbf{K} = W_k \mathbf{Z}$ and $\mathbf{V} = W_v \mathbf{Z}$, where $W_q$, $W_k$ and $W_v$ are learnable parameters.
Given the large number of patch nodes in the graph, we adopt Simplified Linear Attention (SLA)~\cite{guo2024slab} to ensure the computational feasibility of the transformer attention mechanism.
Specifically, for any pair of nodes $v$ and $u$ in the graph, their attention scores are computed as follows:
\begin{equation}
% \small
\mathrm{Sim}_{SLA}\left(Q_{v},K_{u}\right)=\mathrm{ReLU}\left(Q_{v}\right)\mathrm{ReLU}\left(K_{u}\right)^{T},
\end{equation}
\begin{equation}
% \small
\mathrm{GlobalAttn}(\mathbf{Q},\mathbf{K})_{v,u}=\frac{\mathrm{Sim}_{SLA}(Q_{v},K_{u})}{\sum_{u}\mathrm{Sim}_{SLA}(Q_{v},K_{u})}.
\end{equation}
It also benefits from a decoupled computation order by computing $\mathbf{K^T}\mathbf{V}$ first, which allows the computational complexity to be further reduced.
Assume \(\mathbf{Q}\), \(\mathbf{K}\), \(\mathbf{V}\in\mathbb{R}^{N\times D}\), where \(N\) is the number of nodes and \(D\) is the feature dimension. We form the intermediates
\begin{equation}
\mathbf{P}=\mathrm{ReLU}(\mathbf{K})^\top\mathbf{V}\in\mathbb{R}^{D\times D},\quad
\mathbf{y}=\mathrm{ReLU}(\mathbf{K})^\top\mathbf{1}_N\in\mathbb{R}^{D},
\end{equation}
and compute outputs for the intermediate nodes within the block $\mathbf{Z'}$ as
\begin{equation}
\mathbf{Z'}=\frac{\mathrm{ReLU}(\mathbf{Q})\,\mathbf{P}}{\mathrm{ReLU}(\mathbf{Q})\,\mathbf{y}+\varepsilon},
\end{equation}
where \(\varepsilon>0\) is a small constant added for numerical stability. This reduces the computational cost from the standard attention complexity of \(\mathcal{O}(N^2 D)\) to \(\mathcal{O}(N D^2)\), i.e., linear in \(N\) for fixed \(D\), since $N$ is typically larger than $D$.

Specifically, for each node $v \in \mathcal{V}_{L}$, we compute the average attention score of its connected nodes in $\mathcal{V}_{H}$ using the edges in $\mathcal{E}_{L \leftrightarrow H}$ to capture the alignment between them.
The attention maps obtained from $\mathcal{G}_{L}$ and $\mathcal{G}_{H}$ are summed. 
Then, the fusion attention is multiplied by the $\mathbf{V}$ of the Topology Graph branch to incorporate topological features. 
Finally, a prediction head is utilized to compute patch-wise risk scores.

\subsection{Model Self-Interpretability Analysis}
\noindent \textbf{Tissue-Level Interpretability:} 
Our method achieves self-interpretability by explicitly modeling pathological relevance at both patch and slide levels. These patch-level risk scores are then aggregated into a slide-level prediction through mean pooling:
\begin{equation}
% \small
\hat{r}_{\text{slide}} = \frac{1}{M} \sum^{M}_{i=1} r_\text{patch}^{i},
\end{equation}
where \(r_\text{patch}^{i}\) is the risk value associated with the \(i\)-th patch, and \( M \) is the total number of patches within the slide. The patch risk is computed based on the learned representations of local tissue morphology, allowing the model to identify high-risk regions independently. This localized estimation enables a finer-grained assessment of pathological severity. The use of a mean-pooling mechanism is grounded in medical practice. For instance, in prostate cancer, the Gleason score~\cite{gleason1974prediction} is determined by assigning scores to small tissue regions, aligning with aggregating information from localized areas.
This method provides tissue-level interpretability by calculating the risk for each patch, offering a more precise understanding of survival outcomes than traditional attention heatmaps. Lastly, we use the negative log-likelihood survival loss~\cite{dey2022efficient} as survival analysis loss $\mathcal{L}_{risk}$ to supervise the slide risk $\hat{r}_{\text{slide}}$.

\noindent \textbf{Cell-Level Interpretability:} Our method provides a global understanding of features for individual WSIs and cross-cohort, aligning model outputs with biological knowledge.
We utilize cell graphs constructed in \ref{sec:graph_construction} that link model predictions to cell-level interpretability within a set of patches showing each WSI's highest and lowest risk scores.
For every patch $v$, we extract cell statistical feature vectors \( F^{v} \)~\cite{kapse2024si}, such as tumor cell intensity, spatial distribution, and lymphocytic infiltration levels. 
These features, combined with the corresponding risk values, are used to build a Cox~\cite{cox1972regression} proportional hazards regression model, where the coefficients indicate the impact of each feature on risk. The model is formulated as follows:
\begin{equation}\label{Cox}
% \small
r_\text{patch}^{v} = h_0(r) \cdot \exp(\gamma_1 F_1^{v} + \gamma_2 F_2^{v} + \dots + \gamma_p F_p^{v}),
\end{equation}
where \( r_\text{patch}^{v} \) represents the risk given the feature \( F^{v} \), $h_0(r)$ is the control group risk predicted by the output of transformer blocks, and \( \gamma_1, \gamma_2, \dots, \gamma_p \) are the coefficients reflecting the impact of each feature.
This enables global feature interpretation by visualizing the distribution of handcrafted features across the entire cohort. Our approach improves risk prediction without expert annotations, providing an intuitive mapping to biological structures and trusting the rationality of the model for clinical decision-making.

\section{Experiments}
\subsection{Datasets \& Experimental Settings \& Evaluation Metrics}
We validate methods on four public datasets from The Cancer Genome (TCGA)~\cite{weinstein2013cancer}: Breast Invasive Carcinoma (BRCA), Kidney Renal Clear Cell Carcinoma (KIRC), Lung Adenocarcinoma (LUAD), and Stomach Adenocarcinoma (STAD). The data is split into training, validation, and testing sets in a ratio of 60:20:20, and performance is assessed using the average concordance index (C-Index)~\cite{harrell1996multivariable} from 5-fold cross-validation. Statistical significance between the low-risk group and the high-risk group is evaluated using Kaplan-Meier curves with the log-rank test~\cite{mantel1966evaluation}. Models are implemented in PyTorch 1.13.1 on NVIDIA RTX 3090 GPUs, trained for 40 epochs using Adam~\cite{kingma2014adam} optimizer with a constant learning rate of 1e-4 and weight decay of 1e-6.

\subsection{Comparison with State-of-the-Art Methods}
\begin{figure}[H]
\centering
\includegraphics[width=\textwidth]{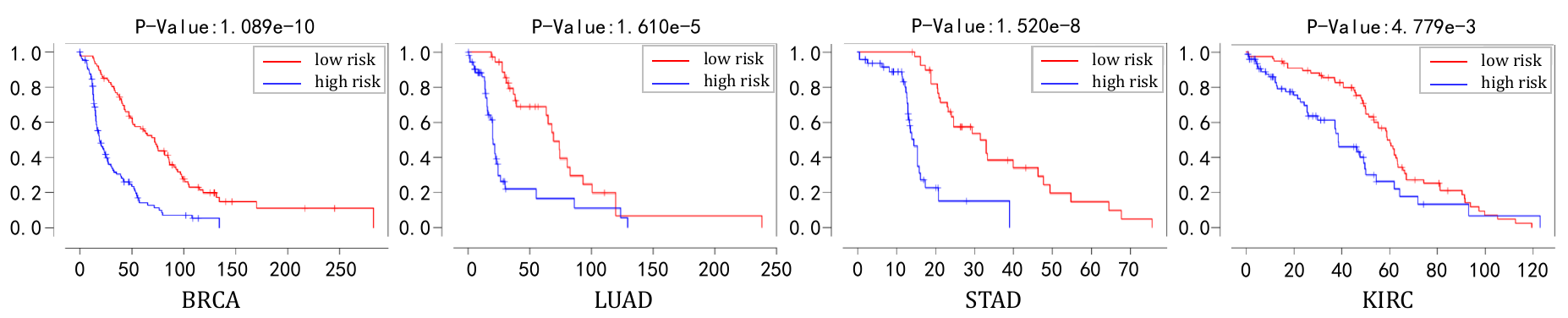}
\caption{Kaplan-Meier curves of IPGPhormer across four datasets. X-axis: survival months, Y-axis: survival probability. }
\label{fig2}
% \vspace{-0.25cm}
\end{figure}

Table~\ref{tab1} compares the C-Index performance across all methods.
Graph-based methods incorporating spatial morphological information outperform attention-based MIL approaches, such as ABMIL~\cite{ilse2018attention} and CSMIL~\cite{deng2024cross}. 
Among graph-based methods, transformer architectures like IPGPhormer better handle long-range information dependencies than GCN-based approaches. 
Our IPGPhormer achieves superior performance across all datasets with a mean C-Index of 0.657, surpassing the second-best model GTP~\cite{zheng2022graph} 0.623 by 3.4\%. This improvement is attributed to its multi-scale feature framework that effectively captures global interactions and fine-grained local logical constraints. 
% The statistical significance of these improvements is further validated by Kaplan-Meier survival curves (Fig.~\ref{fig2}).
Furthermore, in a randomly selected fold, we classify patients into the high-risk group and the low-risk group based on the median predicted risk and plotted Kaplan-Meier survival curves as shown in Fig.~\ref{fig2}. The low p-value of the log-rank test means a significant statistical difference between the low-risk group (red curves) and the high-risk group (blue curves), verifying the statistical significance of the performance improvement of our model.

%%-%%
\begin{table}[htbp]
\centering
\caption{C-Index (mean ± std) over four cancer datasets. The best results are shown in \textbf{bold}, and the second best ones are \underline{underlined}.}\label{tab1}
\resizebox{\textwidth}{!}{
\begin{tabular}{@{}cccccc@{}}
\toprule
\textbf{Model} & \textbf{Multi-Scale} & \textbf{BRCA} & \textbf{KIRC} & \textbf{LUAD} & \textbf{STAD} \\
\midrule

\rowcolor{gray!20} \multicolumn{6}{l}{\textbf{Attention-based}} \\
ABMIL~\cite{ilse2018attention} & \XSolidBrush & $0.570\pm{0.062}$ & $0.645\pm{0.035}$ & $0.584\pm{0.079}$ & $0.605\pm{0.048}$ \\
TransMIL~\cite{shao2021transmil} & \XSolidBrush & $0.606\pm{0.059}$ & $0.651\pm{0.083}$ & $0.555\pm{0.124}$ & $0.599\pm{0.075}$ \\
SurvTRACE~\cite{wang2022survtrace} & \XSolidBrush & $0.563\pm{0.061}$ & $0.581\pm{0.045}$ & $0.556\pm{0.082}$ & $0.575\pm{0.053}$ \\
CSMIL~\cite{deng2024cross} & \Checkmark & $0.543\pm{0.061}$ & $0.541\pm{0.018}$ & $0.571\pm{0.024}$ & $0.510\pm{0.031}$ \\
DSMIL~\cite{li2021dual} & \Checkmark & $0.607\pm{0.038}$ & $0.690\pm{0.081}$ & $0.589\pm{0.116}$ & $0.578\pm{0.056}$ \\
HIPT~\cite{chen2022scaling} & \Checkmark & $0.594\pm{0.046}$ & $0.642\pm{0.028}$ & $0.546\pm{0.041}$ & $0.576\pm{0.089}$ \\
\midrule

\rowcolor{gray!20} \multicolumn{6}{l}{\textbf{Graph-based}} \\
PatchGCN~\cite{chen2021whole} & \XSolidBrush & $0.605\pm{0.048}$ & $0.679\pm{0.025}$ & $0.575\pm{0.079}$ & $0.578\pm{0.058}$ \\
GTP~\cite{zheng2022graph} & \XSolidBrush & $0.598\pm{0.026}$ & $0.697\pm{0.045}$ & $0.575\pm{0.092}$ & \underline{$0.621\pm{0.024}$} \\
HEAT~\cite{chan2023histopathology} & \XSolidBrush & \underline{$0.608\pm{0.037}$} & $0.639\pm{0.087}$ & $0.578\pm{0.104}$ & $0.554\pm{0.079}$ \\
WIKG~\cite{li2024dynamic} & \XSolidBrush & $0.577\pm{0.064}$ & $0.684\pm{0.055}$ & \underline{$0.602\pm{0.043}$} & $0.611\pm{0.041}$ \\
H2MIL~\cite{hou2022h} & \Checkmark & $0.591\pm{0.053}$ & \underline{$0.702\pm{0.075}$} & $0.594\pm{0.114}$ & $0.581\pm{0.047}$ \\
HACT~\cite{pati2022hierarchical} & \Checkmark & $0.564\pm{0.048}$ & $0.683\pm{0.075}$ & $0.534\pm{0.063}$ & $0.559\pm{0.081}$ \\
GRASP~\cite{mirabadi2024grasp} & \Checkmark & $0.597\pm{0.036}$ & $0.675\pm{0.056}$ & $0.575\pm{0.089}$ & $0.592\pm{0.064}$ \\
IPGPhormer(ours) & \Checkmark & \textbf{0.633}$\pm{\textbf{0.030}}$ & \textbf{0.724}$\pm{\textbf{0.058}}$ & \textbf{0.639}$\pm{\textbf{0.063}}$ & \textbf{0.634}$\pm{\textbf{0.037}}$ \\
\bottomrule
\end{tabular}
}
\end{table}
%%-%%

\subsection{Ablation Studies and Hyperparameter Sensitivity}
% We validate the impact of modules in IPGPhormer and hyperparameter sensitivity. 
We explore the selection of the number of nearest neighbors in the graph construction algorithm and present the relevant results in Table~\ref{tab3}. $\mathrm{K}_\mathrm{L}$ and $\mathrm{K}_\mathrm{H}$ denote the K values selected when constructing $\mathcal{G}_{L}$ and $\mathcal{G}_{H}$ using the K-NN algorithm, respectively. By adopting the 8-nearest-neighbor scheme, each node can more comprehensively aggregate information from its neighbors when updating its features. As shown in Table~\ref{tab2}, we further conduct ablation studies to validate the effectiveness of TIE and HIE, which combine morphological and biological information in the Patch-Level Feature Transfer Module. It can be seen that removing them will lead to different degrees of reduction in C-Index, which fully proves that they effectively combine local information and improve the performance of our model. The values in Table~\ref{tab3} and Table~\ref{tab2} are both average C-Index. Fig.~\ref{fig3} shows how the performance varies with different numbers of Transformer layers ($N_L$ and $N_H$) in $\mathcal{G}_{L}$ and $\mathcal{G}_{H}$. The optimal performance is achieved with moderate layer numbers. Considering the balance of performance and computational efficiency, we finally select five self-attention layers in the experiment.
% a和b分别代表在用KNN算法构建C和D时选取的不同K值。 8近邻的构图方法有助于节点更充分的聚合邻居节点信息进行自身特征更新。
\begin{figure}[htbp]
    \centering
    % 左侧表格
    \begin{minipage}{0.49\linewidth}
        \centering
        \begin{table}[H]
          \centering
          \caption{Ablation study on the number of nearest neighbors. \textbf{bold}: the best result, \underline{underlined}: the second best result.}\label{tab3}
          \resizebox{\textwidth}{!}{
            \begin{tabular}{@{}c|cccc@{}}
                \hline
                ($\mathrm{K}_\mathrm{L}$,$\mathrm{K}_\mathrm{H}$) & \textbf{BRCA} & \textbf{KIRC}  & \textbf{LUAD} & \textbf{STAD} \\ 
                \hline
                (4,4) & 0.618 & 0.659 & 0.612 & 0.571 \\ 
                (8,4) & 0.604 & \underline{0.717} & 0.598 & 0.597  \\
                (4,8) & \underline{0.624} & 0.712 & \underline{0.625} & \underline{0.613}\\ 
                (8,8) & \textbf{0.633} & \textbf{0.724} & \textbf{0.639} & \textbf{0.634} \\
                \hline
            \end{tabular}
        }
        \end{table}
    \end{minipage}
    \hfill
    % 右侧图片
    \begin{minipage}{0.49\linewidth}
        \centering
        \begin{table}[H]
            \centering
            \caption{Ablation study on TIE and HIE in Patch-Level Transfer Module. \textbf{bold}: the best result, \underline{underlined}: the second best result.}\label{tab2}
            \resizebox{\textwidth}{!}{
            \begin{tabular}{@{}cc|cccc@{}}
                \hline
                TIE & HIE & \textbf{BRCA} & \textbf{KIRC} & \textbf{LUAD} & \textbf{STAD} \\ 
                \hline
                & & 0.590 & 0.654 & 0.568 & 0.537 \\ 
                \checkmark & & 0.592 & 0.659 & \underline{0.599} & \underline{0.586}  \\
                & \checkmark & \underline{0.624} & \underline{0.673} & 0.592 & 0.541\\ 
                \checkmark & \checkmark & \textbf{0.633} & \textbf{0.724} & \textbf{0.639} & \textbf{0.634} \\
                \hline
            \end{tabular}
        }
        \end{table}
    \end{minipage}
\end{figure}

\begin{figure}[htbp]
    \includegraphics[width=\linewidth]{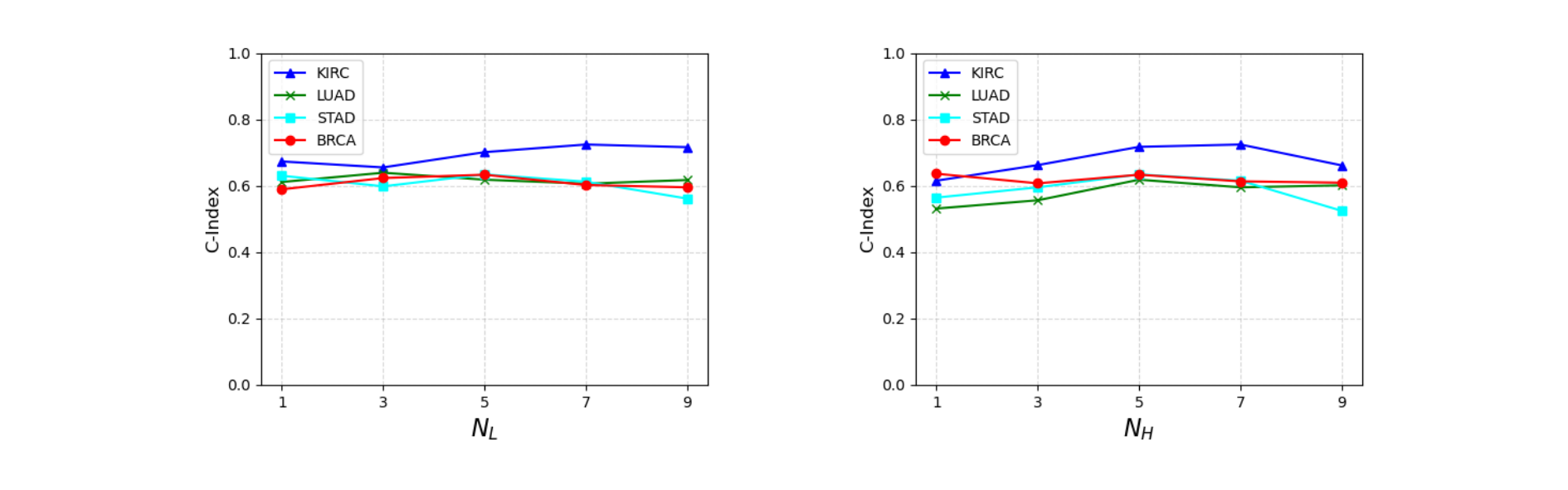}
    \caption{Hyperparameter experiment: Effect of varying the number of Transformer layers.}\label{fig3}
\end{figure}

\subsection{Interpretability}
\begin{figure}[htbp]
\includegraphics[width=\textwidth]{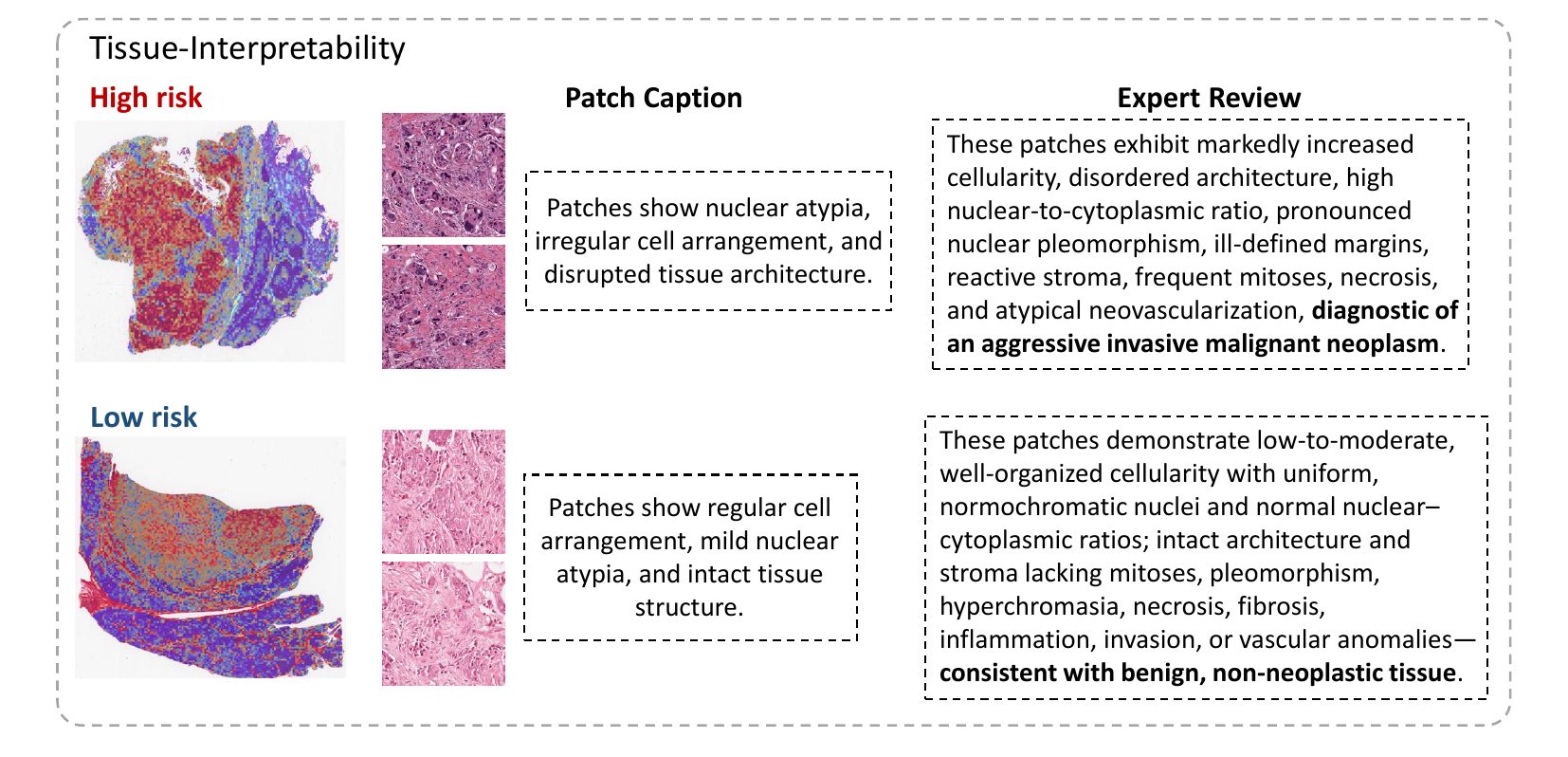}
\caption{The tissue-interpretability approach shows patch-wise risk values aligned with pathology expert assessments.} 
\label{tissue}
\end{figure}

% Fig.~\ref{tissue} illustrates our novel tissue-interpretability approach, differing from traditional attention-based heatmaps. Our model is self-interpretable due to cell-level features used in patch risk estimation, providing each patch with an interpretable risk value. To validate clinical relevance, we analyzed high-risk patches determined by our model from WSIs with varying risk ratings, confirming that their morphological features match pathology experts' review. By examining these features, we clarified the reasons for their high- or low-risk classifications, thus affirming the model's interpretability and accuracy.
Fig.~\ref{tissue} illustrates our novel tissue-interpretability approach, which differs from traditional heatmaps based on attention scores. Thanks to cell-level features involved in patch risk estimation, our model is self-interpretable, as each patch is associated with an interpretable risk value. 
% To validate the clinical relevance of our patch-level risk values, we present some representative high-risk patches, those with the highest risk values calculated by our model, from WSIs with different risk ratings. We find that their internal cell and tissue structural features align with the pathology experts' review. 
To validate the clinical relevance of our patch-level risk values, we present some representative high-risk patches, those with the highest risk values calculated by our model, from WSIs with different risk ratings. Pathology experts' review reveals that the cellular morphology and tissue structure within patches of high predicted risk slides exhibit patterns consistent with high-risk phenotypes, such as increased nuclear pleomorphism, disordered architecture, or disrupted stromal organization. In contrast, patches of low predicted risk slides display more uniform cellular structures and intact tissue organization, align with their predicted low risk classification. 
By analysing their morphological features, we explain the reasons behind their classification as high- or low-risk, further demonstrating the model's interpretability and correctness. 

\begin{figure}[htbp]
\includegraphics[width=\textwidth]{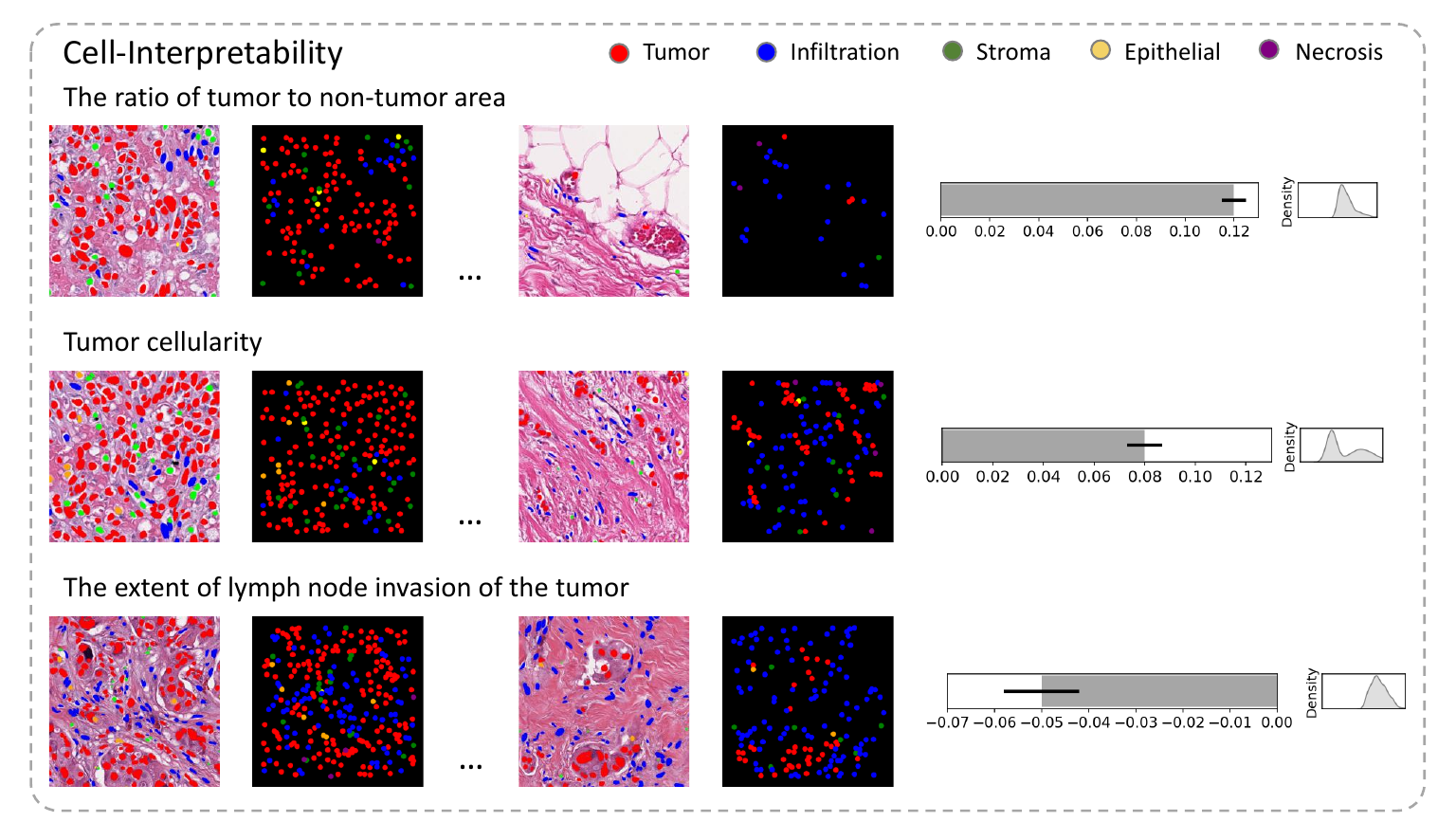}
\caption{
Cross-cohort cell-interpretability analysis aids in the identification of potential biomarkers.} 
\label{fig4}
\end{figure}

In Fig.~\ref{fig4}, we present the cell-interpretability analysis approach cross-cohort. We illustrate how certain statistically significant cellular features influence the patch risk values calculated by Cox regression.
% (see Supporting Materials for details).
The nuclear map identifies the nuclei types and highlights their spatial organization in the patch. The coefficients used to measure the influence of features and the feature distribution are positioned to the right of the nuclear maps. Recall that in Eq.~\ref{Cox}, $\gamma_i$ denotes the contribution of the \(i\)-th feature. The positive contribution represents that the high value of corresponding features will increase the risk score of the patch, while the negative contribution represents that the high value of the corresponding features will reduce the risk prediction.
The feature distribution shows the range of the corresponding normalized features.
For example, the nuclear map reveals densely clustered tumor nuclei in high-risk patches, consistent with a positive coefficient for tumor cell density, while patches enriched in lymphocytes correspond to a negative coefficient for lymphocytic infiltration, reflecting well-established prognostic knowledge that robust immune infiltration is associated with improved outcomes.
% For example, tumor density and the tumor-to-stroma ratio increase the risk, while lymphatic infiltration decreases the risk.
% The global cohort-based cell-level analysis aids in discovering potential prognosis biomarkers. The cell statistical features we extracted can be seen in the supporting materials.
This global cohort-based cell-level analysis not only validates known histopathological biomarkers but also aids in discovering potential cellular signatures with prognostic value. A complete list of extracted cell statistical features and their definitions is provided in the Supporting Materials. 

% \subsection{Computational Complexity}

\section{Conclusion}
In this study, we introduce IPGPhormer, a novel framework designed to predict patch risk and enhance performance and interpretability in survival analysis.
Leveraging a multi-scale graph transformer architecture, IPGPhormer captures intricate features of the tumor microenvironment by integrating local spatial awareness with long-range dependencies. 
Evaluations on four public benchmark datasets demonstrate that IPGPhormer outperforms state-of-the-art methods in predictive accuracy and provides interpretable insights into pathological structures affecting survival outcomes.
In future work, we will integrate additional multimodal data sources and apply IPGPhormer to other pathological tasks, thereby expanding its utility and impact in computational medicine.

\section{Acknowledgements}
This work was supported in part by the National Natural Science Foundation of China under 62031023 \& 62331011; in part by the Shenzhen Science and Technology Project under GXWD20220818170353009.

\bibliography{egbib}

\begin{thebibliography}{37}
\providecommand{\natexlab}[1]{#1}
\providecommand{\url}[1]{\texttt{#1}}
\expandafter\ifx\csname urlstyle\endcsname\relax
  \providecommand{\doi}[1]{doi: #1}\else
  \providecommand{\doi}{doi: \begingroup \urlstyle{rm}\Url}\fi

\bibitem[Aalen(2008)]{aalen2008survival}
Odd~O Aalen.
\newblock \emph{Survival and Event History Analysis: A Process Point of View}.
\newblock Springer-Verlag, 2008.

\bibitem[Chan et~al.(2023)Chan, Cendra, Ma, Yin, and Yu]{chan2023histopathology}
Tsai~Hor Chan, Fernando~Julio Cendra, Lan Ma, Guosheng Yin, and Lequan Yu.
\newblock Histopathology whole slide image analysis with heterogeneous graph representation learning.
\newblock In \emph{Proceedings of the IEEE/CVF conference on computer vision and pattern recognition}, pages 15661--15670, 2023.

\bibitem[Chen et~al.(2021)Chen, Lu, Shaban, Chen, Chen, Williamson, and Mahmood]{chen2021whole}
Richard~J Chen, Ming~Y Lu, Muhammad Shaban, Chengkuan Chen, Tiffany~Y Chen, Drew~FK Williamson, and Faisal Mahmood.
\newblock Whole slide images are 2d point clouds: Context-aware survival prediction using patch-based graph convolutional networks.
\newblock In \emph{Medical Image Computing and Computer Assisted Intervention--MICCAI 2021: 24th International Conference, Strasbourg, France, September 27--October 1, 2021, Proceedings, Part VIII 24}, pages 339--349. Springer, 2021.

\bibitem[Chen et~al.(2022)Chen, Chen, Li, Chen, Trister, Krishnan, and Mahmood]{chen2022scaling}
Richard~J Chen, Chengkuan Chen, Yicong Li, Tiffany~Y Chen, Andrew~D Trister, Rahul~G Krishnan, and Faisal Mahmood.
\newblock Scaling vision transformers to gigapixel images via hierarchical self-supervised learning.
\newblock In \emph{Proceedings of the IEEE/CVF Conference on Computer Vision and Pattern Recognition}, pages 16144--16155, 2022.

\bibitem[Collins and Varmus(2015)]{collins2015new}
Francis~S Collins and Harold Varmus.
\newblock A new initiative on precision medicine.
\newblock \emph{New England journal of medicine}, 372\penalty0 (9):\penalty0 793--795, 2015.

\bibitem[Cox(1972)]{cox1972regression}
David~R Cox.
\newblock Regression models and life-tables.
\newblock \emph{Journal of the Royal Statistical Society: Series B (Methodological)}, 34\penalty0 (2):\penalty0 187--202, 1972.

\bibitem[Deng et~al.(2024)Deng, Cui, Remedios, Bao, Womick, Chiron, Li, Roland, Lau, Liu, et~al.]{deng2024cross}
Ruining Deng, Can Cui, Lucas~W Remedios, Shunxing Bao, R~Michael Womick, Sophie Chiron, Jia Li, Joseph~T Roland, Ken~S Lau, Qi~Liu, et~al.
\newblock Cross-scale multi-instance learning for pathological image diagnosis.
\newblock \emph{Medical image analysis}, 94:\penalty0 103124, 2024.

\bibitem[Dey et~al.(2022{\natexlab{a}})Dey, Zhou, Kiiskinen, Havulinna, Elliott, Karjalainen, Kurki, Qin, FinnGen, Lee, et~al.]{dey2022efficient}
Rounak Dey, Wei Zhou, Tuomo Kiiskinen, Aki Havulinna, Amanda Elliott, Juha Karjalainen, Mitja Kurki, Ashley Qin, FinnGen, Seunggeun Lee, et~al.
\newblock Efficient and accurate frailty model approach for genome-wide survival association analysis in large-scale biobanks.
\newblock \emph{Nature communications}, 13\penalty0 (1):\penalty0 5437, 2022{\natexlab{a}}.

\bibitem[Dey et~al.(2022{\natexlab{b}})Dey, Lipsitz, Cooper, Trinh, Krzywinski, and Altman]{dey2022survival}
Tanujit Dey, Stuart~R Lipsitz, Zara Cooper, Quoc-Dien Trinh, Martin Krzywinski, and Naomi Altman.
\newblock Survival analysis—time-to-event data and censoring.
\newblock \emph{Nat Methods}, 19\penalty0 (8):\penalty0 903, 2022{\natexlab{b}}.

\bibitem[Gamper et~al.(2019)Gamper, Alemi~Koohbanani, Benet, Khuram, and Rajpoot]{gamper2019pannuke}
Jevgenij Gamper, Navid Alemi~Koohbanani, Ksenija Benet, Ali Khuram, and Nasir Rajpoot.
\newblock Pannuke: an open pan-cancer histology dataset for nuclei instance segmentation and classification.
\newblock In \emph{Digital Pathology: 15th European Congress, ECDP 2019, Warwick, UK, April 10--13, 2019, Proceedings 15}, pages 11--19. Springer, 2019.

\bibitem[Gleason and Mellinger(1974)]{gleason1974prediction}
Donald~F Gleason and George~T Mellinger.
\newblock Prediction of prognosis for prostatic adenocarcinoma by combined histological grading and clinical staging.
\newblock \emph{The Journal of urology}, 111\penalty0 (1):\penalty0 58--64, 1974.

\bibitem[Graham et~al.(2019)Graham, Vu, Raza, Azam, Tsang, Kwak, and Rajpoot]{graham2019hover}
Simon Graham, Quoc~Dang Vu, Shan E~Ahmed Raza, Ayesha Azam, Yee~Wah Tsang, Jin~Tae Kwak, and Nasir Rajpoot.
\newblock Hover-net: Simultaneous segmentation and classification of nuclei in multi-tissue histology images.
\newblock \emph{Medical image analysis}, 58:\penalty0 101563, 2019.

\bibitem[Guo et~al.(2024)Guo, Chen, Tang, and Wang]{guo2024slab}
Jialong Guo, Xinghao Chen, Yehui Tang, and Yunhe Wang.
\newblock Slab: Efficient transformers with simplified linear attention and progressive re-parameterized batch normalization.
\newblock \emph{arXiv preprint arXiv:2405.11582}, 2024.

\bibitem[Harrell~Jr et~al.(1996)Harrell~Jr, Lee, and Mark]{harrell1996multivariable}
Frank~E Harrell~Jr, Kerry~L Lee, and Daniel~B Mark.
\newblock Multivariable prognostic models: issues in developing models, evaluating assumptions and adequacy, and measuring and reducing errors.
\newblock \emph{Statistics in medicine}, 15\penalty0 (4):\penalty0 361--387, 1996.

\bibitem[Hou et~al.(2022)Hou, Yu, Lin, Huang, Yu, Qin, and Wang]{hou2022h}
Wentai Hou, Lequan Yu, Chengxuan Lin, Helong Huang, Rongshan Yu, Jing Qin, and Liansheng Wang.
\newblock H\^{} 2-mil: exploring hierarchical representation with heterogeneous multiple instance learning for whole slide image analysis.
\newblock In \emph{Proceedings of the AAAI conference on artificial intelligence}, volume~36, pages 933--941, 2022.

\bibitem[Ilse et~al.(2018)Ilse, Tomczak, and Welling]{ilse2018attention}
Maximilian Ilse, Jakub Tomczak, and Max Welling.
\newblock Attention-based deep multiple instance learning.
\newblock In \emph{International conference on machine learning}, pages 2127--2136. PMLR, 2018.

\bibitem[Jackson et~al.(2020)Jackson, Fischer, Zanotelli, Ali, Mechera, Soysal, Moch, Muenst, Varga, Weber, et~al.]{jackson2020single}
Hartland~W Jackson, Jana~R Fischer, Vito~RT Zanotelli, H~Raza Ali, Robert Mechera, Savas~D Soysal, Holger Moch, Simone Muenst, Zsuzsanna Varga, Walter~P Weber, et~al.
\newblock The single-cell pathology landscape of breast cancer.
\newblock \emph{Nature}, 578\penalty0 (7796):\penalty0 615--620, 2020.

\bibitem[Kapse et~al.(2024)Kapse, Pati, Das, Zhang, Chen, Vakalopoulou, Saltz, Samaras, Gupta, and Prasanna]{kapse2024si}
Saarthak Kapse, Pushpak Pati, Srijan Das, Jingwei Zhang, Chao Chen, Maria Vakalopoulou, Joel Saltz, Dimitris Samaras, Rajarsi~R Gupta, and Prateek Prasanna.
\newblock Si-mil: Taming deep mil for self-interpretability in gigapixel histopathology.
\newblock In \emph{Proceedings of the IEEE/CVF Conference on Computer Vision and Pattern Recognition}, pages 11226--11237, 2024.

\bibitem[Kingma and Ba(2014)]{kingma2014adam}
Diederik~P Kingma and Jimmy Ba.
\newblock Adam: A method for stochastic optimization.
\newblock \emph{arXiv preprint arXiv:1412.6980}, 2014.

\bibitem[Lee et~al.(2022)Lee, Park, Oh, Shin, Sun, Jung, Lee, Kim, Chung, Moon, et~al.]{TEA}
Yongju Lee, Jeong~Hwan Park, Sohee Oh, Kyoungseob Shin, Jiyu Sun, Minsun Jung, Cheol Lee, Hyojin Kim, Jin-Haeng Chung, Kyung~Chul Moon, et~al.
\newblock Derivation of prognostic contextual histopathological features from whole-slide images of tumours via graph deep learning.
\newblock \emph{Nature Biomedical Engineering}, pages 1--15, 2022.

\bibitem[Li et~al.(2021)Li, Li, and Eliceiri]{li2021dual}
Bin Li, Yin Li, and Kevin~W Eliceiri.
\newblock Dual-stream multiple instance learning network for whole slide image classification with self-supervised contrastive learning.
\newblock In \emph{Proceedings of the IEEE/CVF conference on computer vision and pattern recognition}, pages 14318--14328, 2021.

\bibitem[Li et~al.(2024)Li, Chen, Chu, Sun, Guan, Han, and He]{li2024dynamic}
Jiawen Li, Yuxuan Chen, Hongbo Chu, Qiehe Sun, Tian Guan, Anjia Han, and Yonghong He.
\newblock Dynamic graph representation with knowledge-aware attention for histopathology whole slide image analysis.
\newblock In \emph{Proceedings of the IEEE/CVF Conference on Computer Vision and Pattern Recognition}, pages 11323--11332, 2024.

\bibitem[Lin et~al.(2021)Lin, Lan, and Li]{lin2021generative}
Wanyu Lin, Hao Lan, and Baochun Li.
\newblock Generative causal explanations for graph neural networks.
\newblock In \emph{International Conference on Machine Learning}, pages 6666--6679. PMLR, 2021.

\bibitem[Liu and Kurc(2022)]{liu2022deep}
Huidong Liu and Tahsin Kurc.
\newblock Deep learning for survival analysis in breast cancer with whole slide image data.
\newblock \emph{Bioinformatics}, 38\penalty0 (14):\penalty0 3629--3637, 2022.

\bibitem[Liu et~al.(2024)Liu, Ji, Ye, and Fu]{liu2024advmil}
Pei Liu, Luping Ji, Feng Ye, and Bo~Fu.
\newblock Advmil: Adversarial multiple instance learning for the survival analysis on whole-slide images.
\newblock \emph{Medical Image Analysis}, 91:\penalty0 103020, 2024.

\bibitem[Lu et~al.(2021)Lu, Williamson, Chen, Chen, Barbieri, and Mahmood]{lu2021data}
Ming~Y Lu, Drew~FK Williamson, Tiffany~Y Chen, Richard~J Chen, Matteo Barbieri, and Faisal Mahmood.
\newblock Data-efficient and weakly supervised computational pathology on whole-slide images.
\newblock \emph{Nature biomedical engineering}, 5\penalty0 (6):\penalty0 555--570, 2021.

\bibitem[Mantel et~al.(1966)]{mantel1966evaluation}
Nathan Mantel et~al.
\newblock Evaluation of survival data and two new rank order statistics arising in its consideration.
\newblock \emph{Cancer Chemother Rep}, 50\penalty0 (3):\penalty0 163--170, 1966.

\bibitem[Mirabadi et~al.(2024)Mirabadi, Archibald, Darbandsari, Contreras-Sanz, Nakhli, Asadi, Zhang, Gilks, Black, Wang, et~al.]{mirabadi2024grasp}
Ali~Khajegili Mirabadi, Graham Archibald, Amirali Darbandsari, Alberto Contreras-Sanz, Ramin~Ebrahim Nakhli, Maryam Asadi, Allen Zhang, C~Blake Gilks, Peter Black, Gang Wang, et~al.
\newblock Grasp: graph-structured pyramidal whole slide image representation.
\newblock \emph{arXiv preprint arXiv:2402.03592}, 2024.

\bibitem[Pati et~al.(2022)Pati, Jaume, Foncubierta-Rodriguez, Feroce, Anniciello, Scognamiglio, Brancati, Fiche, Dubruc, Riccio, et~al.]{pati2022hierarchical}
Pushpak Pati, Guillaume Jaume, Antonio Foncubierta-Rodriguez, Florinda Feroce, Anna~Maria Anniciello, Giosue Scognamiglio, Nadia Brancati, Maryse Fiche, Estelle Dubruc, Daniel Riccio, et~al.
\newblock Hierarchical graph representations in digital pathology.
\newblock \emph{Medical image analysis}, 75:\penalty0 102264, 2022.

\bibitem[Shao et~al.(2021)Shao, Bian, Chen, Wang, Zhang, Ji, et~al.]{shao2021transmil}
Zhuchen Shao, Hao Bian, Yang Chen, Yifeng Wang, Jian Zhang, Xiangyang Ji, et~al.
\newblock Transmil: Transformer based correlated multiple instance learning for whole slide image classification.
\newblock \emph{Advances in neural information processing systems}, 34:\penalty0 2136--2147, 2021.

\bibitem[Vaswani et~al.(2017)Vaswani, Shazeer, Parmar, Uszkoreit, Jones, Gomez, Kaiser, and Polosukhin]{vaswani2017attention}
Ashish Vaswani, Noam Shazeer, Niki Parmar, Jakob Uszkoreit, Llion Jones, Aidan~N Gomez, {\L}ukasz Kaiser, and Illia Polosukhin.
\newblock Attention is all you need.
\newblock \emph{Advances in neural information processing systems}, 30, 2017.

\bibitem[Veli{\v{c}}kovi{\'c} et~al.(2017)Veli{\v{c}}kovi{\'c}, Cucurull, Casanova, Romero, Lio, and Bengio]{velivckovic2017graph}
Petar Veli{\v{c}}kovi{\'c}, Guillem Cucurull, Arantxa Casanova, Adriana Romero, Pietro Lio, and Yoshua Bengio.
\newblock Graph attention networks.
\newblock \emph{arXiv preprint arXiv:1710.10903}, 2017.

\bibitem[Wang et~al.(2022)Wang, Yang, Zhang, Wang, Zhang, Yang, Huang, and Han]{wang2022transformer}
Xiyue Wang, Sen Yang, Jun Zhang, Minghui Wang, Jing Zhang, Wei Yang, Junzhou Huang, and Xiao Han.
\newblock Transformer-based unsupervised contrastive learning for histopathological image classification.
\newblock \emph{Medical image analysis}, 81:\penalty0 102559, 2022.

\bibitem[Wang et~al.(2023)Wang, Gao, Yi, Zhang, Zhang, Zhang, Li{\`o}, Bain, Bassed, Li, et~al.]{Surformer}
Zhikang Wang, Qian Gao, Xiaoping Yi, Xinyu Zhang, Yiwen Zhang, Daokun Zhang, Pietro Li{\`o}, Chris Bain, Richard Bassed, Shanshan Li, et~al.
\newblock Surformer: An interpretable pattern-perceptive survival transformer for cancer survival prediction from histopathology whole slide images.
\newblock \emph{Computer Methods and Programs in Biomedicine}, 241:\penalty0 107733, 2023.

\bibitem[Wang and Sun(2022)]{wang2022survtrace}
Zifeng Wang and Jimeng Sun.
\newblock Survtrace: Transformers for survival analysis with competing events.
\newblock In \emph{Proceedings of the 13th ACM international conference on bioinformatics, computational biology and health informatics}, pages 1--9, 2022.

\bibitem[Weinstein et~al.(2013)Weinstein, Collisson, Mills, Shaw, Ozenberger, Ellrott, Shmulevich, Sander, and Stuart]{weinstein2013cancer}
John~N Weinstein, Eric~A Collisson, Gordon~B Mills, Kenna~R Shaw, Brad~A Ozenberger, Kyle Ellrott, Ilya Shmulevich, Chris Sander, and Joshua~M Stuart.
\newblock The cancer genome atlas pan-cancer analysis project.
\newblock \emph{Nature genetics}, 45\penalty0 (10):\penalty0 1113--1120, 2013.

\bibitem[Zheng et~al.(2022)Zheng, Gindra, Green, Burks, Betke, Beane, and Kolachalama]{zheng2022graph}
Yi~Zheng, Rushin~H Gindra, Emily~J Green, Eric~J Burks, Margrit Betke, Jennifer~E Beane, and Vijaya~B Kolachalama.
\newblock A graph-transformer for whole slide image classification.
\newblock \emph{IEEE transactions on medical imaging}, 41\penalty0 (11):\penalty0 3003--3015, 2022.

\end{thebibliography}
\end{document}